\documentclass[letterpaper, 10 pt, conference]{ieeeconf}  % Comment this line out if you need a4paper
\usepackage[utf8]{inputenc}

\IEEEoverridecommandlockouts                              % This command is only needed if 
\usepackage{amsmath} % assumes amsmath package installed
\usepackage{amssymb}  % assumes amsmath package installed
\usepackage{graphicx}
\usepackage{amsmath}
\usepackage{amssymb}
\usepackage{booktabs}
\usepackage{lipsum} % for dummy text
\usepackage[absolute,overlay]{textpos}
\usepackage{multirow}
\usepackage{url}
\usepackage{xcolor}
\usepackage{color}
\usepackage{pifont}
\usepackage{hyperref}
\usepackage{wrapfig}
\usepackage{multirow}
\usepackage{tabularx}
\usepackage{caption}
\usepackage{mathtools}
\usepackage{bm}
\usepackage{pifont}
\usepackage{makecell}
\usepackage{siunitx} 
\usepackage{bbm}
\usepackage{colortbl}
\usepackage{dsfont}
\usepackage{bbm}
\usepackage{placeins}
\usepackage{wrapfig}
\usepackage[top=0.75in, bottom=0.78in, left=0.75in, right=0.75in]{geometry}
\usepackage{microtype}
\usepackage{balance}

\newcommand{\cmark}{\ding{51}}
\newcommand{\xmark}{\ding{55}}

\hypersetup{pdfstartview=FitH,
            colorlinks=true,
            linkcolor=red,
            anchorcolor=blue,
            citecolor=black
            }
 
\definecolor{tablecolor}{HTML}{ccf2f5} 
\newcommand{\dd}[2]{$#1\scriptstyle{\pm#2}$}
\newcommand{\ddbf}[2]{$\mathbf{#1\scriptstyle{\pm#2}}$}

\title{\LARGE \bf
ISS Policy : Scalable Diffusion Policy with Implicit Scene Supervision
}

\author{Wenlong Xia$^{*}$, Jinhao Zhang$^{*}$, Ce Zhang, Yaojia Wang, Huizhe Li, Yichen Lai, Yude Li, Youmin Gong and Jie Mei$\textsuperscript{‡}$
\thanks{* W. Xia and J. Zhang contribute equally to this work.}
\thanks{‡ Corresponding author}}

\begin{document}

\maketitle
\thispagestyle{empty}
\pagestyle{empty}
\raggedbottom

% %%%%%%%%%%%%%%%%%% 修改开始：插入跨栏图片 %%%%%%%%%%%%%%%%%%
% \begin{strip}
%     \centering
%     % 调整图片上方的垂直间距，避免紧贴作者栏
%     \vspace{-1.6cm} 
    
%     \includegraphics[width=\textwidth]{pics/radar.pdf}
    
%     % 注意：在非浮动环境中使用 \captionof{figure}{...} 而不是 \caption{...}
% \captionof{figure}{\small \textbf{ISS Policy} is a novel 3D visuomotor diffusion-based policy that generates continuous action sequences from point cloud inputs. It exhibits strong scalability and learning efficiency, achieving state-of-the-art performance on challenging simulation benchmarks and demonstrating strong robustness in real-world experiments.\label{fig:system}}

%     % 调整图片下方的垂直间距，避免紧贴摘要
%     \vspace{-0.2cm} 
% \end{strip}
% %%%%%%%%%%%%%%%%%% 修改结束 %%%%%%%%%%%%%%%%%%

\begin{abstract}
Vision-based imitation learning has enabled impressive robotic manipulation skills, but action imitation alone provides limited supervision of how robot behavior changes the surrounding 3D scene. To address this challenge, we introduce \emph{Implicit Scene Supervision (ISS) Policy}, a 3D visuomotor DiT-based diffusion policy that predicts sequences of continuous actions from point cloud observations. We extend DiT with an implicit scene supervision module that predicts robot motion from the generated actions and uses this motion to forecast future scene geometry. By modeling how actions translate into robot motion, this module encourages the policy to capture the geometric consequences of robot--environment interactions. We further separate motion regression from scene-level policy supervision and balance prediction errors across different scene-change magnitudes. These designs provide dynamics-aware geometric supervision using only expert demonstrations, without requiring additional annotations or auxiliary predictors at inference. Notably, ISS Policy achieves state-of-the-art performance across single-arm manipulation, dexterous hand manipulation, and bimanual manipulation tasks. Real-world experiments further demonstrate state-of-the-art performance in robotic manipulation. The resulting framework retains the scalable DiT backbone and the standard diffusion-policy control interface. Code and videos will be released.

\end{abstract}

%%%%%%%%%%%%%%%%%%%%%%%%%%%%%%%%%%%%%%%%%%%%%%%%%%%%%%%%%%%%%%%%%%%%%%%%%%%%%%%%
\section{INTRODUCTION}
Visuomotor imitation learning maps visual observations and robot states to actions using expert demonstrations~\cite{levine2016end,zhang2018deep,mandlekar2021matters}. Diffusion policies provide an expressive model of continuous action sequences~\cite{chi2025diffusion}, while point-cloud observations supply the geometry needed for spatially precise manipulation~\cite{ze20243d,shridhar2023perceiver}. Combining compact 3D representations with a diffusion transformer (DiT)~\cite{peebles2023scalable} offers a practical foundation for learning control from demonstrations. A remaining question is how to use the temporal structure of those demonstrations to supervise the policy beyond matching expert actions.

Future-scene prediction offers one such signal: an action representation should carry information about how the observed geometry will evolve. Geometric prediction tasks have been explored as auxiliary supervision for visuomotor learning~\cite{ze2023visual}. In a DiT policy, a future-latent head can condition on the predicted action sequence and current point-cloud features, using a later demonstration observation as its target. This avoids reconstructing a dense scene or requiring object-pose and contact annotations. However, the interface between the action predictor and the scene predictor determines what the auxiliary task actually teaches.

An action specifies the intended behavior, whereas the resulting robot motion also depends on the current robot state, controller response, saturation, and contact. For example, the same translational action may move a freely moving end effector or produce little displacement against an obstacle. A gripper action likewise expresses closing intent even after contact prevents further closure. Conditioning a scene predictor directly on actions leaves it to learn both how actions produce robot motion and how that motion changes the scene. Moreover, direct motion regression through a shared policy can compete with action imitation, and an unnormalized future-feature loss can underweight small scene changes relative to large ones. These issues motivate separating the variables and gradient paths used for control and predictive supervision.

We introduce \textbf{Implicit Scene Supervision (ISS) Policy}. The policy generates actions in the demonstrated action space. During training, a supervised GRU motion predictor estimates the resulting end-effector displacement from actions and robot-state context. Its prediction is combined with the original gripper intent to form a representation that conditions a residual predictor of future point-cloud features. 

We train this pathway with asymmetric gradient routing. Direct motion regression receives detached actions and state features, so it trains the motion predictor without directly pulling the policy toward a local motion objective. Future-scene losses use predicted motion without detaching it and can therefore shape the policy through the motion predictor. In addition to absolute future-feature regression, a change-balanced loss normalizes each sample's error by its target-change magnitude, with a lower bound for nearly static samples. An optional late curriculum reduces the two predictive losses during policy refinement while retaining direct motion supervision. 

Our contributions are:
\begin{itemize}
\item A scalable 3D diffusion policy, ISS Policy, connecting action generation to future scene prediction through supervised robot motion prediction.
\item A training-time implicit scene supervision scheme that improves policy performance and learning efficiency, while introducing no additional inference cost since the auxiliary predictors are disabled at deployment.
\item Extensive experiments across multiple simulation benchmarks demonstrate state-of-the-art performance, while real-world evaluations further validate the effectiveness and practical applicability of the proposed design.
\end{itemize}
\section{RELATED WORKS}
\subsection{Visuomotor Imitation Learning}
Imitation learning from expert demonstrations has emerged as a powerful paradigm for acquiring robotic manipulation skills~\cite{osa2018algorithmic,mandlekar2021matters}. Numerous works have explored end-to-end imitation learning directly from 2D visual observations, mapping RGB images to actions via convolutional or transformer-based policies~\cite{chi2025diffusion,florence2022implicit,haldar2023teach,reuss2024multimodal,liu2024rdt,gong2025carp}. While these methods prove simple and effective on standard benchmarks, their inherent lack of depth and 3D geometry often results in brittle behavior during fine-grained, multi-object, and long-horizon manipulation tasks~\cite{zhu2024point,wang2024rise}.

This realization has spurred a shift towards 3D-aware policies leveraging explicit geometric representations~\cite{gervet2023act3d,ze20243d,goyal2023rvt,ke20243d,shridhar2023perceiver,chen2025g3flow}. While methods based on dense voxel grids~\cite{shridhar2023perceiver,goyal2023rvt} or continuous 3D feature fields~\cite{gervet2023act3d,ke20243d,chen2025g3flow} offer rich spatial details, they are often compute-intensive, limiting training scalability and inference speed. Conversely, a more efficient alternative is to use a sparse point-cloud encoder, as introduced in DP3~\cite{ze20243d}. Building upon this efficiency, we use single-view point clouds to condition an action diffusion policy. Our auxiliary supervision connects actions to future scene features through a supervised robot motion predictor, with separate gradient routes for motion regression and scene prediction.

\subsection{Diffusion Models for Robotics}
%%%%%%%%%%%%%%%%%%%%%%

\begin{figure*}[!t]
    \centering
    \includegraphics[width=\textwidth]{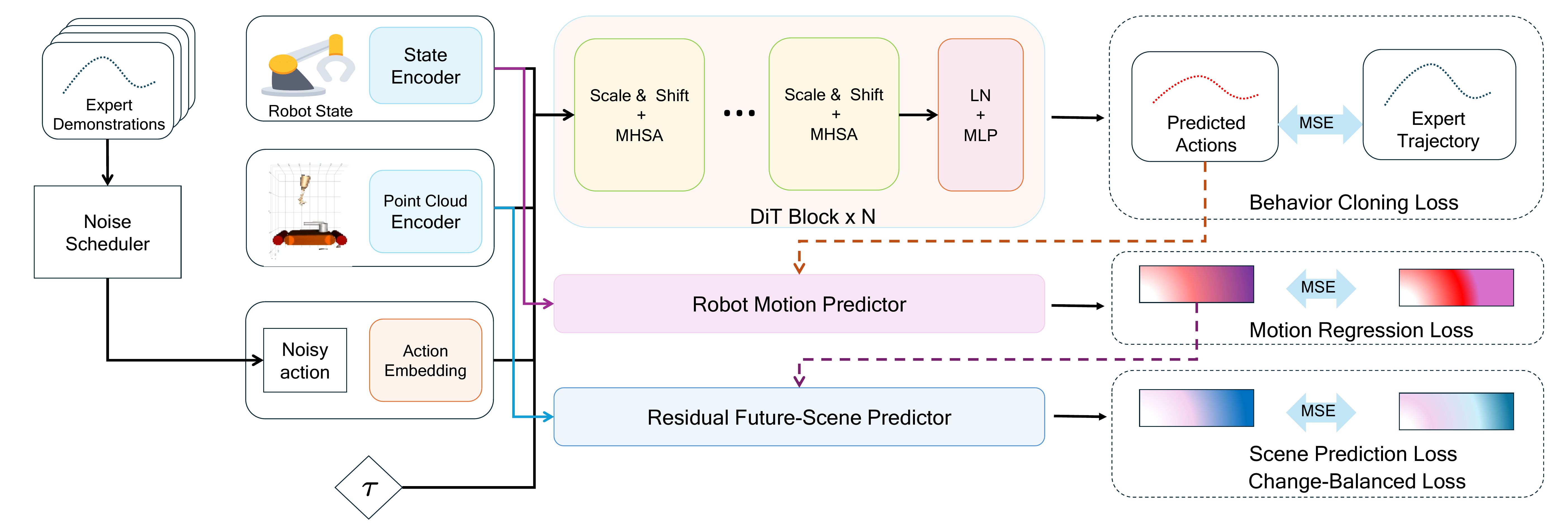}
    \caption{\small \textbf{Model Architecture of ISS Policy.}
    Point-cloud and robot-state observations condition a DiT-based policy that denoises action trajectories.
    The implicit scene supervision branch provides an auxiliary training signal from skip-step future point-cloud embeddings.}
    \label{fig:overview}
\end{figure*}
%%%%%%%%%%%%%%%%%%%%%%
Diffusion models are a class of generative models that transform random noise into complex data through a gradual denoising process, and were popularized for high-fidelity image synthesis~\cite{ho2020denoising,song2020denoising,song2020score,rombach2022high}.  

As diffusion models matured beyond image synthesis, the community began applying them to decision making and robotic control. Diffusion Policy~\cite{chi2025diffusion} first established the paradigm of representing policies as conditional diffusion processes. This paradigm has subsequently seen widespread adoption across a broad spectrum of tasks, including general manipulation~\cite{ke20243d,xian2023chaineddiffuser,cao2024mamba,yan2025m}, grasping~\cite{wu2023learning,urain2023se,barad2024graspldm,weng2024dexdiffuser}, and navigation~\cite{sridhar2024nomad,cai2025navdp,samavi2025sicnav}. To further enhance expressiveness and scalability, RDT~\cite{liu2024rdt} introduced the Diffusion Transformer (DiT)~\cite{peebles2023scalable} backbone to robotic manipulation, demonstrating the superiority of DiT in handling heterogeneous robotic data. Leveraging the superior scalability and computational efficiency of DiT, we design a specialized policy architecture tailored for robotic manipulation with point clouds input, to generate precise action sequences.

\section{PRELIMINARIES}
\subsection{Diffusion Models}
 The basic idea of Denoising Diffusion Probabilistic Models (DDPM)~\cite{ho2020denoising} is to construct a simple forward diffusion process that progressively destroys structure in the data by adding noise, and then learn a reverse process that reconstructs clean data from noisy inputs.

\noindent\textbf{Forward Diffusion.} The forward process defines a sequence $\{\mathbf{x}_t\}_{t=1}^T$ by repeatedly adding Gaussian noise according to a Markov chain:
\begin{equation}
    q(\mathbf{x}_t | \mathbf{x}_{t-1}) = \mathcal{N}(\mathbf{x}_t; \sqrt{1 - \beta_t}\mathbf{x}_{t-1}, \beta_t\mathbf{I}), \quad t=1, \dots, T,
\end{equation}
where $\{\beta_t\}_{t=1}^T$ is a predefined variance schedule with $\beta_t \in (0,1)$. Because the forward process is linear and Gaussian, one can also sample any intermediate state in closed form:
\begin{equation}
    q(\mathbf{x}_t | \mathbf{x}_0) = \mathcal{N}(\mathbf{x}_t; \sqrt{\bar{\alpha}_t}\mathbf{x}_0, (1 - \bar{\alpha}_t)\mathbf{I}),
\end{equation}
where $\alpha_t = 1- \beta_t$ and $\bar{\alpha}_t = \prod_{s=1}^t \alpha_s.$
As $t$ increases, $x_t$ gradually converges to standard Gaussian.

\noindent\textbf{Reverse Denoising.} The reverse process begins by sampling $\mathbf{x}_T \sim \mathcal{N}(0, \mathbf{I})$ and iteratively recover the original data $x_0$. The exact reverse conditionals $q(\mathbf{x}_{t-1}|\mathbf{x}_{t})$ are intractable, so they are approximated by a parameterized Gaussian
\begin{equation}
    p_\theta(\mathbf{x}_{t-1} | \mathbf{x}_t) = \mathcal{N}(\mathbf{x}_{t-1}; \boldsymbol{\mu}_\theta(\mathbf{x}_t, t), \boldsymbol{\Sigma}_\theta(\mathbf{x}_t, t)),
\end{equation}
where a neural network estimates either the mean $\mu_\theta$ or equivalently the noise contained in $\mathbf{x}_t$.
In DDPM, the variance is typically fixed as a function of the schedule, and the network $\epsilon_\theta(\mathbf{x}_t,t)$ is trained to predict the true noise $\epsilon$ used to construct $\mathbf{x}_t$. The objective reduces to a simple mean-squared error
\begin{equation}
    \mathcal{L}_{\text{ddpm}} = \mathbb{E}_{t, \mathbf{x}_0, \boldsymbol{\epsilon}} \left[ \left\| \boldsymbol{\epsilon} - \boldsymbol{\epsilon}_\theta(\sqrt{\bar{\alpha}_t}\mathbf{x}_0 + \sqrt{1 - \bar{\alpha}_t}\boldsymbol{\epsilon}, t) \right\|_2^2 \right],
\end{equation}
which can be interpreted as a variational bound on the negative log-likelihood. 

In this work, we use the same forward noising process with a clean-sample prediction parameterization: the network estimates $\mathbf{x}_0$ directly instead of predicting the injected noise. We use DDIM-style sampling~\cite{song2020denoising} for action generation, as specified in the method below.

\section{METHOD}
ISS Policy augments action diffusion with a supervised robot motion predictor and a residual future-scene predictor (Fig.~\ref{fig:overview}). We first define the executable policy, then describe the auxiliary variables, gradient routing, and training objective.

\subsection{Problem Formulation}
An expert trajectory records observations $o_t=(P_t,s_t)$ and actions $a_t$, where $P_t$ is a single-view point cloud, $s_t$ denotes the robot state, and $a_t\in\mathbb{R}^{d_a}$ represents the robot control command. Depending on the embodiment, $a_t$ may contain end-effector motion, joint-level control, and gripper commands. Importantly, the commanded motion in $a_t$ represents the intended robot movement and is distinct from the actual robot motion measured from successive states.

\begin{figure}[tbp]
    \centering
    \includegraphics[width=0.95\linewidth]{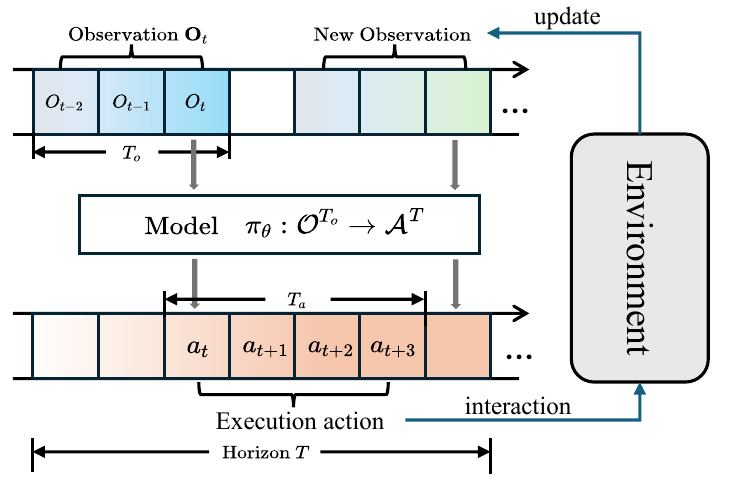}
    \caption{\textbf{Policy Formulation.}}
    \label{fig:action}
\end{figure}

Following Diffusion Policy~\cite{chi2025diffusion}, we use an observation window of length $T_o$, an action prediction window of length $H$, and execute $T_a$ actions before replanning (Fig.~\ref{fig:action}). The aligned action window is $A_t=(a_\tau,\ldots,a_{\tau+H-1})$, where $\tau=t-T_o+1$. Execution starts at the latest observed timestep $t$, corresponding to token $T_o-1$ in zero-based indexing. 

\subsection{Action Diffusion}
Following DP3~\cite{ze20243d}, a PointNet-style encoder uses pointwise MLPs and max pooling to produce $z_t=E_{\mathrm{pc}}(P_t)\in\mathbb R^{d_z}$. An MLP encodes the robot state as $c_t=E_s(s_t)$. We concatenate the observation-window features into $Z_t=[z_\tau,\ldots,z_t]$ and $C_t=[c_\tau,\ldots,c_t]$. Given a noisy action sequence $A_t^k$ at diffusion step $k$, the DiT estimates the clean actions:
\begin{equation}
    \hat A_t^0=D_\theta(A_t^k,k,Z_t,C_t).
    \label{eq:action}
\end{equation}
Noisy action embeddings are first fused with projected point-cloud context and temporal embeddings, and the resulting action sequence is then processed by self-attention layers. Meanwhile, projected point-cloud, state, and diffusion-step features are jointly injected into each diffusion transformer block through AdaLN-Zero~\cite{peebles2023scalable}. The denoiser predicts clean actions using the masked behavior-cloning objective
\begin{equation}
    \mathcal L_{\mathrm{BC}}
    =\mathbb E_{A_t,k,\epsilon}
    \left[\left\|M\odot(\hat A_t^0-A_t)\right\|_2^2\right],
    \label{eq:bc}
\end{equation}
where $M$ selects the supervised trajectory entries; the loss is averaged over batch and trajectory dimensions. The auxiliary branches described below do not change the action dimension, prediction window, or receding-horizon execution.

\subsection{Robot Motion Prediction}
To supervise robot motion prediction, ISS Policy derives end-effector displacement targets from successive recorded states:
\begin{equation}
    m_j=p^{\mathrm{eef}}_{j+1}-p^{\mathrm{eef}}_j.
\end{equation}
Here $p^{\mathrm{eef}}_j$ is the end-effector position available in $s_j$. Motion normalization uses within-episode transitions, excluding reset discontinuities. We retain the gripper action to represent opening or closing intent rather than supervising aperture change, which is strongly affected by contact and saturation.

During training, the motion predictor receives either demonstrated or predicted actions, selected independently for each sample:
\begin{equation}
    \tilde A_t=bA_t+(1-b)\hat A_t^0,\quad
    b\sim\operatorname{Bernoulli}(p(u)),
    \label{eq:mix}
\end{equation}
\begin{equation}
    p(u)=p_0+(p_f-p_0)\min(u/U_{\mathrm{TF}},1),
\end{equation}
where $u$ is the optimization step. We use $p_0=0.5$, $p_f=0$, and $U_{\mathrm{TF}}=10{,}000$. This schedule introduces predicted actions gradually.

A GRU motion predictor $R_\psi$ embeds each action and initializes its hidden state from $C_t$. It predicts normalized end-effector displacements, which are combined with gripper intent to form motion-and-gripper tokens:
\begin{equation}
    \hat M_t=R_\psi(\tilde A_t,C_t),\qquad
    v_{\tau+j}=[\hat M_{t,j},\,\tilde g_{\tau+j}]\in\mathbb R^4,
    \label{eq:motion}
\end{equation}
where $\tilde g$ is the gripper component of the selected action. The token thus combines a supervised estimate of physical translation with the original gripper intent.

\subsection{Asymmetric Gradient Routing}
We distinguish direct motion regression from the route used for scene prediction. A second forward pass through the same motion predictor uses detached inputs:
\begin{equation}
    \begin{aligned}
    \hat M_t^{\mathrm{det}}
        &=R_\psi(\operatorname{sg}(\tilde A_t),
                  \operatorname{sg}(C_t)),\\
    \mathcal L_{\mathrm{motion}}
        &=\frac{1}{H-1}\sum_{j=0}^{H-2}
          \left\|\hat M_{t,j}^{\mathrm{det}}-\bar m_{\tau+j}\right\|_2^2,
    \end{aligned}
    \label{eq:motion_loss}
\end{equation}
where $\operatorname{sg}(\cdot)$ denotes stop-gradient and $\bar m_j$ is the normalized motion target. The last motion token lacks a successor observation in the sampled window and is excluded. This loss updates $\psi$ only, so its relatively strong regression signal cannot directly update predicted actions or the state encoder.

The scene predictor uses the motion-and-gripper tokens from Eq.~\eqref{eq:motion} without detaching them. For a future offset $K=2$, it predicts a residual relative to the latest point-cloud latent:
\begin{equation}
    \hat z_{t+K}
    =\operatorname{sg}(z_t)
     +F_\omega(v_{t:t+K-1},\operatorname{sg}(Z_t)).
    \label{eq:future}
\end{equation}
The head projects point-cloud context and the $K$ motion-and-gripper token embeddings, uses the latter to condition scale, shift, and gating, and outputs a latent residual. The scene prediction loss is
\begin{equation}
    \mathcal L_{\mathrm{scene}}
    =\frac{1}{B d_z}\sum_{i=1}^B
      \left\|\hat z_{t+K,i}
                  -\operatorname{sg}(z_{t+K,i})\right\|_2^2.
    \label{eq:scene_loss}
\end{equation}
Current point-cloud inputs and future targets are detached at the scene head, while the observation encoders remain trainable through the BC objective along the main policy pathway. This gradient routing prevents the local motion supervision from directly updating the policy, while preserving the scene-prediction supervision.

\subsection{Change-Balanced Scene Supervision}
Absolute feature regression is sensitive to the scale of each sample's scene change. We additionally normalize prediction error relative to the target change, so small changes can contribute without requiring a large absolute latent displacement. Define
\begin{equation}
    \begin{aligned}
    \Delta z_i
        &=\operatorname{sg}(z_{t+K,i})-\operatorname{sg}(z_{t,i}),\\
    \widehat{\Delta z}_i
        &=\hat z_{t+K,i}-\operatorname{sg}(z_{t,i}),\\
    r_i&=\max\left(\sqrt{\|\Delta z_i\|_2^2/d_z},\,\varepsilon_c\right).
    \end{aligned}
    \label{eq:change_scale}
\end{equation}
With $\varepsilon_c=0.05$, the change-balanced objective is
\begin{equation}
    \mathcal L_{\mathrm{change}}
      =\frac{1}{B d_z}\sum_{i=1}^B
       \left\|\frac{\widehat{\Delta z}_i-\Delta z_i}{r_i}\right\|_2^2.
    \label{eq:change_loss}
\end{equation}
The denominator is detached, and its lower bound limits amplification for nearly static samples. This is a sample-dependent reweighting of future-feature error: it shares the target and prediction of $\mathcal L_{\mathrm{scene}}$, rather than introducing another prediction head. The two terms retain absolute prediction accuracy while balancing samples with different change magnitudes.

\subsection{Training Objective}
The motion, scene prediction, and change-balanced losses jointly constitute the implicit scene supervision (ISS) loss:
\begin{equation}
    \begin{aligned}
    \mathcal L_{\mathrm{ISS}}={}&\lambda_m\mathcal L_{\mathrm{motion}}\\
      &+\lambda_s\mathcal L_{\mathrm{scene}}
         +\lambda_c\mathcal L_{\mathrm{change}},
    \end{aligned}
    \label{eq:iss_loss}
\end{equation}
where $\lambda_m$, $\lambda_s$, and $\lambda_c$ weight the three components. The complete objective is
\begin{equation}
    \mathcal L=\mathcal L_{\mathrm{BC}}
      +\lambda_{\mathrm{ISS}}\mathcal L_{\mathrm{ISS}},
    \label{eq:total_loss}
\end{equation}
where $\lambda_{\mathrm{ISS}}$ controls the overall strength of implicit scene supervision. We use component weights $\lambda_m=0.1$, $\lambda_s=0.01$, and $\lambda_c=0.001$. 

The optional late auxiliary curriculum scales the two scene-loss weights $\lambda_s$ and $\lambda_c$ within $\mathcal L_{\mathrm{ISS}}$ by $s(u)$ during action-policy refinement:
\begin{equation}
    s(u)=
    \begin{cases}
      1, & u\le U_s,\\
      1+(\rho-1)\dfrac{u-U_s}{U_f-U_s}, & U_s<u<U_f,\\
      \rho, & u\ge U_f.
    \end{cases}
    \label{eq:curriculum}
\end{equation}

 At deployment, DDIM sampling uses 10 denoising steps to generate actions; the next three actions are executed before replanning. The auxiliary heads are disabled during inference, making implicit scene supervision inference-cost-free.

\begin{table*}[!t]
\centering
\caption{\textbf{Quantitative comparisons of different baselines in Adroit and MetaWorld.} 
We compare our ISS Policy with BCRNN, IBC, Diffusion Policy, DP3, and Mamba Policy on simulation tasks in terms of SR$_5$. 
Unavailable results for BCRNN and IBC are denoted by ``--''. Results reproduced with the official code repository are denoted by *.}
\label{tab:main_exp}
% resizebox 包裹整个 tabular
\resizebox{1\textwidth}{!}{%
    \begin{tabular}{l ccc cccccc c}
    \toprule
    & \multicolumn{3}{c}{\textbf{Adroit}} & \multicolumn{6}{c}{\textbf{MetaWorld}} & \\
    \cmidrule(lr){2-4} \cmidrule(lr){5-10} % 关键：只在对应列下方画横线，(lr)表示左右留白
    Method 
    & Hammer & Door & Pen 
    & Assembly & Disassemble & Stick-Push & Reach-Wall & Stick-Pull & Pick-Place-Wall 
    & \textbf{Average} \\
    \midrule
    
    BCRNN
    & \dd{0}{0} & \dd{0}{0} & \dd{9}{3} 
    & \dd{3}{4} & \dd{32}{12} & \dd{45}{11}  
    & -- & -- & -- 
    & -- \\
    
    IBC
    & \dd{0}{0} & \dd{0}{0} & \dd{9}{2}  
    & \dd{0}{0} & \dd{1}{1} & \dd{16}{2}  
    & -- & -- & -- 
    & -- \\
    
    Diffusion Policy
    & \dd{48}{17} & \dd{50}{5} & \dd{25}{4} 
    & \dd{15}{1} & \dd{43}{7} & \dd{63}{3}
    & \dd{59}{7} & \dd{11}{2} & \dd{5}{1}
    & 35.4 \\
    
    DP3*
    & \ddbf{100.0}{0.0} & \dd{62}{4} & \dd{43.7}{5.3}  
    & \dd{99.6}{0.4} & \dd{75.0}{3.0} & \ddbf{100.0}{0.0}      
    & \dd{70.7}{6.7} & \dd{74.3}{4.3} & \dd{82.7}{5}
    & 78.7 \\
    
    Mamba Policy*
    & \ddbf{100.0}{0.0} & \dd{63.8}{7.8} & \dd{42.3}{5.2}  
    & \ddbf{100.0}{0.0} & \dd{81.3}{3.7} & \ddbf{100.0}{0.0}
    & \dd{78.3}{4.3} & \dd{79.3}{1.3} & \dd{88.0}{2.0}
    & 81.4 \\
    
    \midrule 
    \textbf{ISS Policy (Ours)} 
    & \ddbf{100.0}{0.0} & \ddbf{72.0}{2.0} & \ddbf{52.3}{2.7} 
    & \ddbf{100.0}{0.0} & \ddbf{91.7}{4.7} & \ddbf{100.0}{0.0}
    & \ddbf{88.0}{0.7} & \ddbf{85.3}{4.7} & \ddbf{90.7}{1.9}
    & \textbf{86.7}  \\
    \bottomrule
    \end{tabular}%
}
\end{table*}

\begin{table*}[t]
      \centering
      \caption{\textbf{Quantitative comparisons of different baselines in RoboTwin2.0.} Each task is evaluated across 100 randomly generated scenes.}
      \label{tab:robotwin2_iss_selected}
      \resizebox{\textwidth}{!}{
          \begin{tabular}{ccccccc}
          \toprule
          ~ & Beat Block Hammer & Click Alarmclock & Click Bell
            & Dump Bin Bigbin & Handover Block & Move Can Pot \\
          \midrule
          DP
            & 42.0 & 61.0 & 54.0 & 49.0 & 10.0 & 39.0 \\
          $\pi_0$
            & 43.0 & 63.0 & 44.0 & 83.0 & 45.0 & 58.0 \\
          DP3
            & \textbf{72.0} & 77.0 & 90.0 & 85.0 & 70.0 & 70.0 \\
          \rowcolor{gray!10}
          \textbf{ISS-Policy (Ours)}
            & \textbf{72.0}
            & \textbf{88.0}
            & \textbf{100.0}
            & \textbf{88.0}
            & \textbf{75.0}
            & \textbf{86.0} \\

          \toprule
          ~ & Move Playingcard Away & Open Laptop & Place Bread Skillet
            & Place Can Basket & Place Cans Plasticbox & Place Container Plate \\
          \midrule
          DP
            & 47.0 & 49.0 & 11.0 & 18.0 & 40.0 & 41.0 \\
          $\pi_0$
            & 53.0 & \textbf{85.0} & \textbf{23.0}
            & 41.0 & 34.0 & 88.0 \\
          DP3
            & 68.0 & 82.0 & 19.0 & 67.0 & 48.0 & 86.0 \\
          \rowcolor{gray!10}
          \textbf{ISS-Policy (Ours)}
            & \textbf{70.0}
            & 84.0
            & 20.0
            & \textbf{72.0}
            & \textbf{72.0}
            & \textbf{90.0} \\

          \toprule
          ~ & Place Empty Cup & Place Object Stand & Press Stapler
            & Stack Bowls Two & Stamp Seal & Turn Switch \\
          \midrule
          DP
            & 37.0 & 22.0 & 6.0 & 61.0 & 2.0 & 36.0 \\
          $\pi_0$
            & 37.0 & 36.0 & 62.0 & \textbf{91.0} & 3.0 & 27.0 \\
          DP3
            & 65.0 & 60.0 & 69.0 & 83.0 & 18.0 & 46.0 \\
          \rowcolor{gray!10}
          \textbf{ISS-Policy (Ours)}
            & \textbf{76.0}
            & \textbf{71.0}
            & \textbf{70.0}
            & 84.0
            & \textbf{23.0}
            & \textbf{53.0} \\
          \bottomrule
          \end{tabular}
      }
  \end{table*}
\begin{figure*}[!t]
    \centering
    \includegraphics[width=\textwidth]{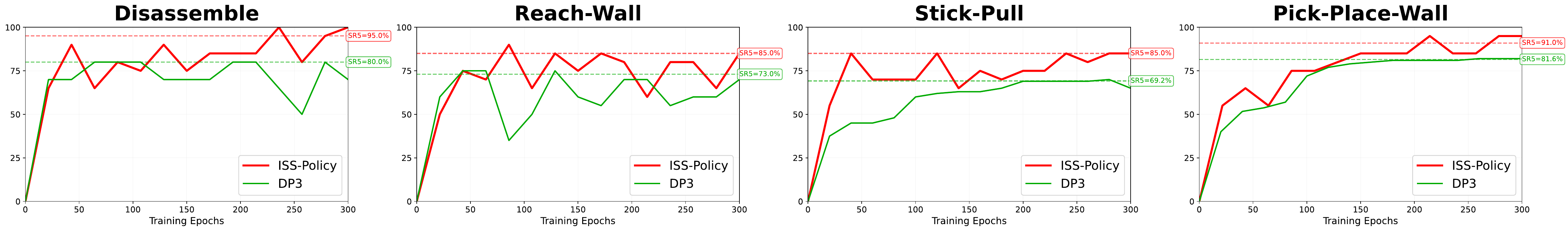}
    \caption{\textbf{Learning Efficiency and Stability. } The dashed horizontal lines denote the SR$_5$ metric. Across all tasks, ISS Policy
    achieves higher success rates, higher learning efficiency, and better asymptotic
    performance.}

    \label{fig:training_curve}
\end{figure*}

\section{EXPERIMENTS}

%picture:3d obsevations   
\noindent\textbf{Simulation Benchmark.}
Simulation benchmarks play a crucial role in evaluating visuomotor policies before deployment on physical robots. We evaluate our approach on simulated manipulation tasks from the Adroit~\cite{rajeswaran2017learning}, MetaWorld~\cite{yu2020meta}, and RoboTwin2.0~\cite{chen2025robotwin} benchmarks. Adroit focuses on dexterous hand manipulation with high-dimensional continuous control, offering a challenging testbed for precise, long-horizon skills. MetaWorld provides a diverse set of single-arm manipulation tasks (e.g., pushing, picking, and placing) categorized into multiple difficulty levels following prior work~\cite{seo2023masked}. RoboTwin2.0 further extends our evaluation to bimanual manipulation, providing a diverse suite of dual-arm collaborative tasks with strong domain randomization over objects and scene configurations.
% The 3D observations are visualized in Figure ~\ref{fig:point_cloud}.
% %%%%%%%%%%%%%%%%%%%%%%
% %%%%%%%%%%%%%%%%%%%%%%
% \begin{figure}[!ht]
%     \centering
%     \includegraphics[width=\linewidth]{pics/point_cloud.pdf}
%     \caption{\textbf{Visualization of 3D point-cloud observations in simulation environments.} We show representative 3D point clouds from Adroit and MetaWorld.}
%     \label{fig:point_cloud}
%     \vspace{-2mm}
% \end{figure}
% %%%%%%%%%%%%%%%%%%%%%%% 

%%%%%%%%%%%%%%%%%%%%%%%

\noindent\textbf{Expert Demonstrations.}
For the Adroit tasks, we collect successful trajectories using reinforcement learning (RL) agents trained with VRL3~\cite{wang2022vrl3}, which has been shown to produce strong expert policies in dexterous manipulation domains. For MetaWorld tasks, we follow the standard protocol and use built-in scripted policies to generate expert demonstrations. For RoboTwin2.0, we follow the official protocol and train each task using 50 expert demonstrations. In all cases, we only retain successful rollouts and use them as expert demonstrations for imitation learning. 

\noindent\textbf{Baselines.}
Our goal is to highlight the advantages of our proposed architecture, so our main baseline is the 3D-based policy DP3~\cite{ze20243d}, which represents the current SOTA in point cloud conditioned policies. In addition, we include the 2D-based policy DP~\cite{chi2025diffusion}, IBC~\cite{florence2022implicit}  and BCRNN~\cite{mandlekar2021matters}, 3D-based policy Mamba Policy~\cite{cao2024mamba} and VLA policy $\pi_0$~\cite{black2024pi0} . All baselines are trained on the same set of expert demonstrations, and we match the image and depth resolutions across all 2D and 3D variants for fair comparison. 

\noindent\textbf{Evaluation Metrics.}
For Adroit and MetaWorld, we train three runs with random seeds {0,1,2} and evaluate each model every 200 epochs using 20 rollouts per task. We report \(\mathrm{SR}_5\), defined as the average of the top-five success rates during training, with the mean and standard deviation computed across three seeds. For RoboTwin 2.0, we follow the official protocol and report the average success rate over 100 randomly generated scenes with 100 different random seeds for each task.

\noindent\textbf{Experiment Settings.}
We set the prediction horizon to $T=4$, with observation steps $T_o=2$ and action steps $T_a=3$. Following DP3~\cite{ze20243d}, We adopt a DDIM noise scheduler, using 100 diffusion steps for training and 10 steps for inference. Optimization is performed with AdamW, with an initial learning rate of $1 \times 10^{-4}$ and a cosine decay schedule. Both actions and robot states are normalized to $[-1, 1]$ before training, and actions are rescaled back for execution. All models are trained for 3000 epochs with a batch size of 128. Our model and all baselines are trained and evaluated on a single NVIDIA RTX 5880 GPU.

%%

%%
%%%%%%%%%%%%%%%%%%%%%%%%%%%%%%%%%%%%%%%%%%%%%%%%%%%%%%%%%%%%%%%%%%%%%%%%%%%%%%%%%%%%%%%%%%%%%%%%%
%%
%%%%%%%%%%%%%%%%%%%%%%

%%%%%%%%%%%%%%%%%%%%%%%```

\subsection{Comparisons with the State-of-the-Arts}
\label{sec:sota}

As shown in Tab.~\ref{tab:main_exp}, we compare ISS Policy with state-of-the-art visuomotor baselines on Adroit and MetaWorld in terms of $\text{SR}_5$. Classical behavior-cloning baselines such as BCRNN and IBC perform poorly (average $\text{SR}_5 \leq 15\%$), suggesting that traditional behavior cloning with a unimodal regression objective tends to suffer from mode averaging. Among all methods, ISS Policy attains the highest average success rate, consistently outperforming both 2D and 3D diffusion baselines.

These results suggest that ISS Policy benefits from dynamics-aware geometric supervision, which improves perception--action coupling and action learning from 3D observations.

\subsection{Scalability Evaluation}

We investigate how ISS Policy scales along two dimensions: the capacity of the DiT-based action head and the amount of demonstration data.

\begin{figure}[tbp]
\centering

% ===== 上半部分：表格 =====
\begin{minipage}{0.95\linewidth}
\renewcommand{\arraystretch}{1.1}
\centering
\captionof{table}{\textbf{DiT configurations in the scalability study.} 
We consider three variants (Small, Medium, Large) with increasing depth, width, and parameters.}
\label{tab:dit_scales}
\resizebox{0.99\linewidth}{!}{%
\begin{tabular}{l|cccc}
\hline\hline
\multicolumn{1}{c|}{\textbf{Model Size}} 
& \multicolumn{1}{c}{$n_{\text{head}}$} 
& \multicolumn{1}{c}{Depth} 
& \multicolumn{1}{c}{Hidden dim} 
& \multicolumn{1}{c}{Params (M)} \\
\hline
Small  & 8  & 8  & 640  & 44.98  \\
Medium & 8 & 12  & 784  & 92.78  \\
Large  & 8 & 16 & 1024 & 214.61 \\
\hline\hline
\end{tabular}}
\end{minipage}
\par\medskip

% ===== 下半部分：图片 =====
\begin{minipage}{0.9\linewidth}
\centering
\includegraphics[width=\linewidth]{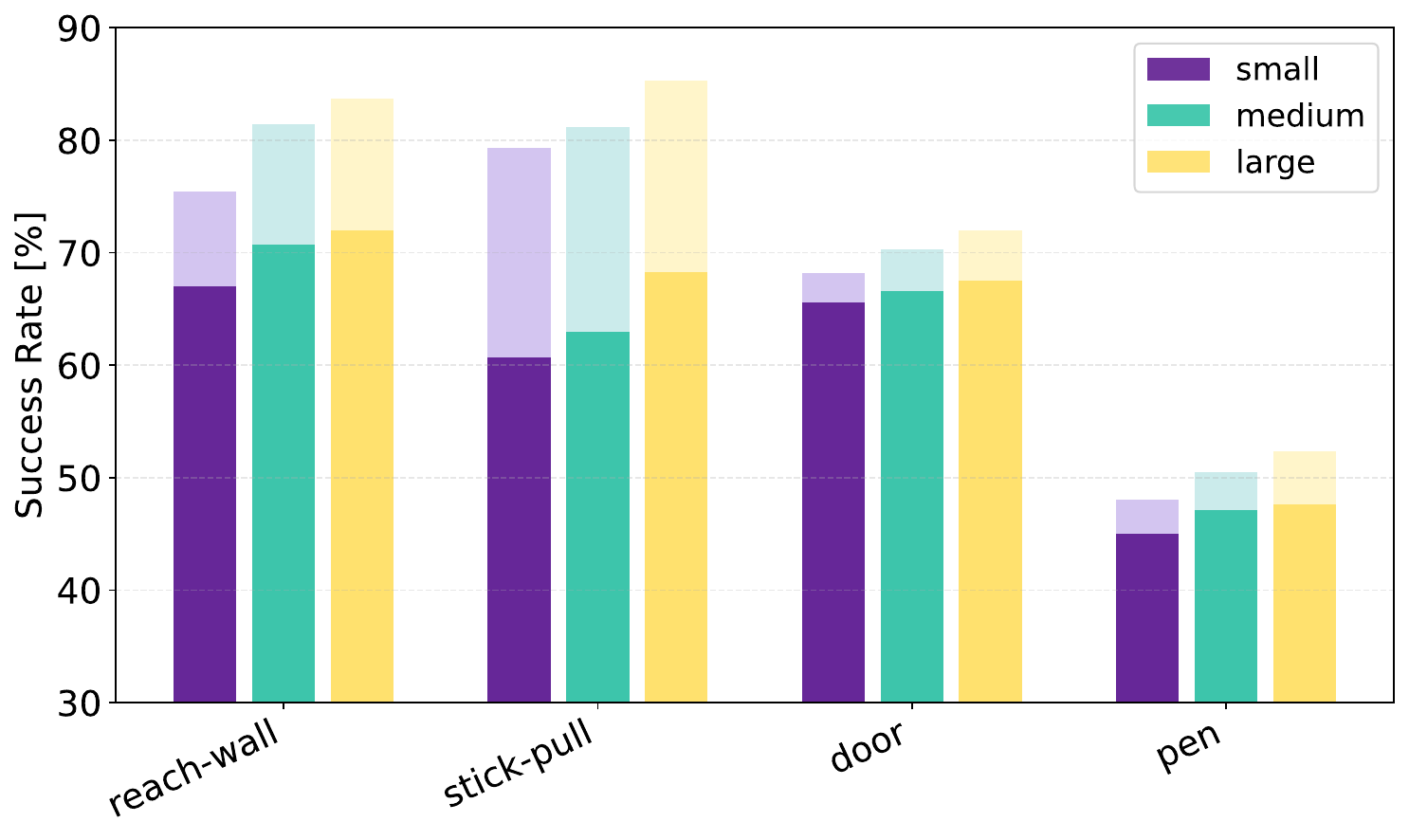}
\captionof{figure}{\textbf{Scalability Study of ISS Policy with DiT model size.}
We visualize $\text{SR}_5$ on four tasks using three DiT policy heads from Table~\ref{tab:dit_scales}, with light bars (w/ ISS) and dark bars (w/o ISS).}

\label{fig:scaling}
\end{minipage}
\end{figure}

\noindent\textbf{Scaling with model capacity.}
Table~\ref{tab:dit_scales} and Fig.~\ref{fig:scaling} sweep the DiT policy-head capacity from Small to Large. Across four tasks, larger DiT heads consistently achieve higher $\text{SR}_5$, with performance improving monotonically as model capacity increases. These results suggest that our architecture continues to benefit from increased policy capacity, rather than exhibiting early saturation at small scales.

\begin{table}[tbp]
\centering
\caption{\textbf{Scalability Study of demonstration numbers. }We evaluate $\text{SR}_5$ of ISS Policy under different numbers of demonstrations on three tasks.} 
\label{tab:iss_demo}
\resizebox{0.99\columnwidth}{!}{%
\begin{tabular}{c|ccc|c} 
\toprule
\# Demos & Door & Pen & Reach-Wall & Average \\
\midrule
$10$  & \dd{72.0}{2.0} & \dd{52.3}{2.7} & \dd{83.7}{1.3} & 69.3 \\
$30$  & \dd{73.8}{6.8} & \dd{69.0}{6.0} & \dd{85.0}{4.0} & 75.9 \\
$50$  & \dd{80.8}{0.8} & \dd{83.3}{2.8} & \dd{87.0}{3.0} & 83.7 \\
\bottomrule
\end{tabular}}
\end{table}

\noindent\textbf{Scaling with training data.}
Table~\ref{tab:iss_demo} sweeps the number of expert demonstrations while keeping the architecture fixed. ISS Policy attains strong performance with only 10 demonstrations and continues to improve as the dataset size increases, with the most pronounced gains on the more challenging Pen task. These trends suggest that ISS Policy is data-efficient in the low-data regime and continues to benefit from additional demonstrations.

%%%%%%%%%%%%%%%%%%%%%%%%%%%%%%%%%%%%%%%%%%%%%%%%%%%%%%%%%%
\subsection{Learning Efficiency and Stability}
We further compare the training curves of ISS Policy against DP3. As shown in Fig.~\ref{fig:training_curve}, ISS Policy consistently attains high performance much earlier in training than DP3. The success rate of ISS Policy increases sharply in the early training phase, whereas DP3 improves gradually and often requires substantially more epochs to reach comparable performance.

To quantitatively assess training stability, we examine the standard deviations reported in Tab.~\ref{tab:main_exp}. Our method exhibits the lowest variance while also achieving the highest success rate. These results suggest that ISS Policy not only converges faster but also produces more consistent performance across different seeds.
% indicating that the DiT-based architecture together with implicit scene supervision yields a more efficient policy that can exploit demonstration data more effectively.

%%%%%%%%%%%%%%%%%%%%%%%%%%%%%%%%%%%%%%%%%%%%%%%%%%%%%%%%%%%%%%%%%%%%%%%%%%

%%%%%%%%%%%%%%%%%%%%%%%%%%%%%%%%%%%%%%%%%%%%%%%%%%%%%%%%%%%%%%%%%%%%%%%%%%%

%%%%%%%%%%%%%%%%%%%%%%%%%%%%%%%%%%%%%%%%%%%%%%%%%%%%%%%%%%%%
\begin{table}[tbp]
\centering
\caption{\textbf{Ablation on Designed Module.} We evaluate Conditional Fusion, ISS Loss ($\mathcal L_{\mathrm{ISS}}$), and Schedule Sampling, reporting the average success rates on Adroit and MetaWorld.}
\label{tab:cond_iss_ablation}
% resizebox 必须包裹住整个 tabular 环境
\resizebox{0.95\linewidth}{!}{%
    \begin{tabular}{ccc cc} % 学术表格通常建议去除竖线 |
    \toprule
    \multicolumn{3}{c}{\textbf{Modules}} & \multicolumn{2}{c}{\textbf{Success Rate}} \\
    \cmidrule(lr){1-3} \cmidrule(lr){4-5} % 在 1-3 列和 4-5 列下方画局部横线
    Cond. Fusion & ISS Loss & Schedule Samp. & Adroit & MetaWorld \\
    \midrule
    \xmark & \xmark & \xmark & 69.30 & 85.00 \\
    \cmark & \xmark & \xmark & 70.77 & 87.78 \\
    \cmark & \cmark & \xmark & 70.83 & 89.33   \\
    \cmark & \cmark & \cmark & 74.77 & 91.68 \\
    \bottomrule
    \end{tabular}%
}
\end{table}

\subsection{Ablation Study}
\noindent\textbf{Impact of Designed Module.}
Table~\ref{tab:cond_iss_ablation} shows that conditional fusion provides modest gains by better grounding action queries in the 3D point-cloud context. However, adding ISS on top of fusion delivers only marginal improvements, because the auxiliary loss is supervised solely under predicted actions. In contrast, enabling schedule sampling on top of ISS stabilizes this auxiliary supervision under model-predicted actions and mitigates exposure bias, thereby achieving the best performance. Together, these modules are complementary: conditional fusion refines perception–action coupling; ISS shapes future-aware context representations; and schedule sampling makes this supervision effective under the model’s own action distribution, thereby improving robustness when rolling out the learned policy.

\begin{figure}[!htb]
    \centering
    \begin{minipage}[b]{0.49\linewidth}
        \centering
        \includegraphics[width=\linewidth]{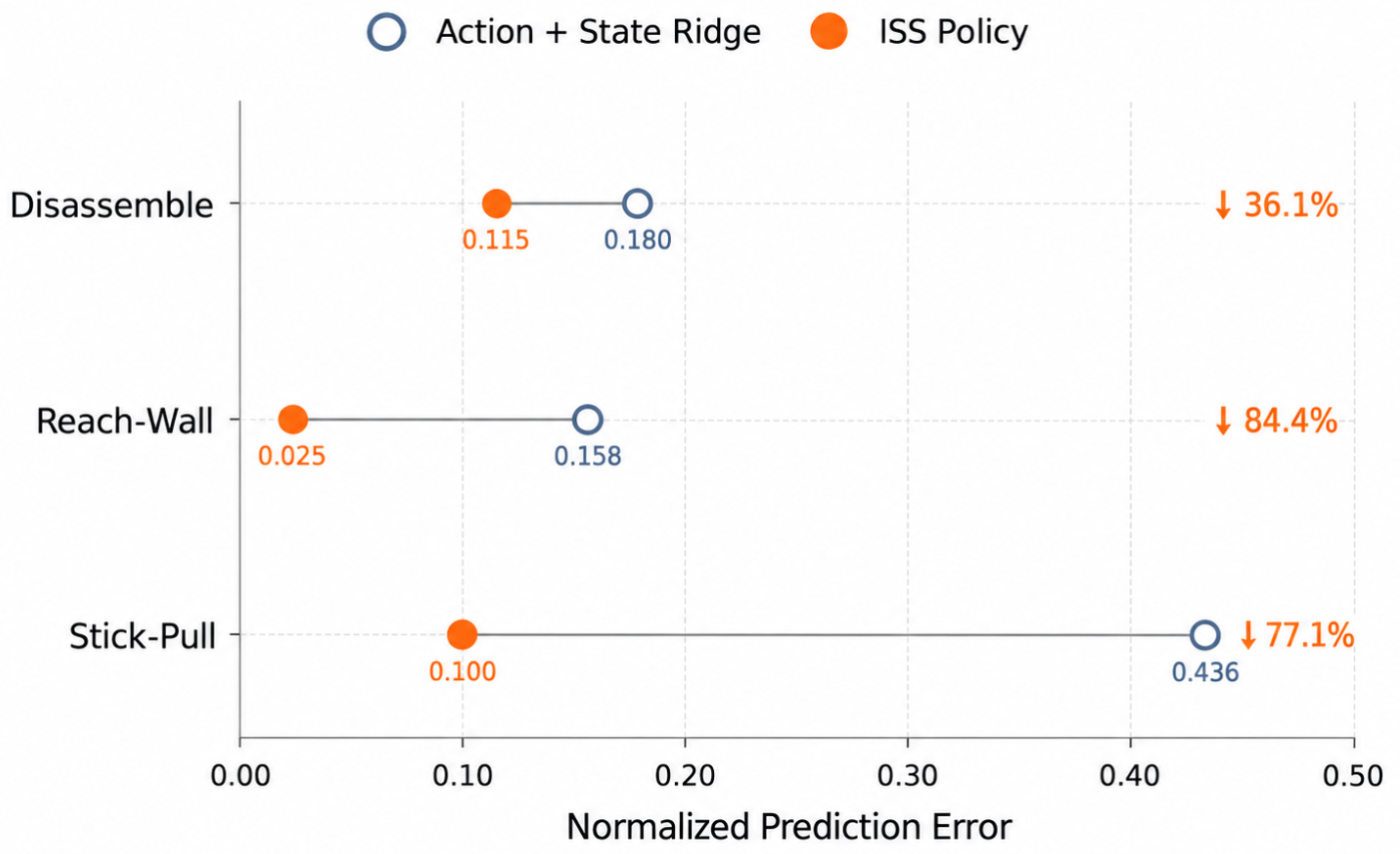}
        \par\smallskip
        {\footnotesize (a) Robot motion prediction}
    \end{minipage}\hfill
    \begin{minipage}[b]{0.49\linewidth}
        \centering
        \includegraphics[width=\linewidth]{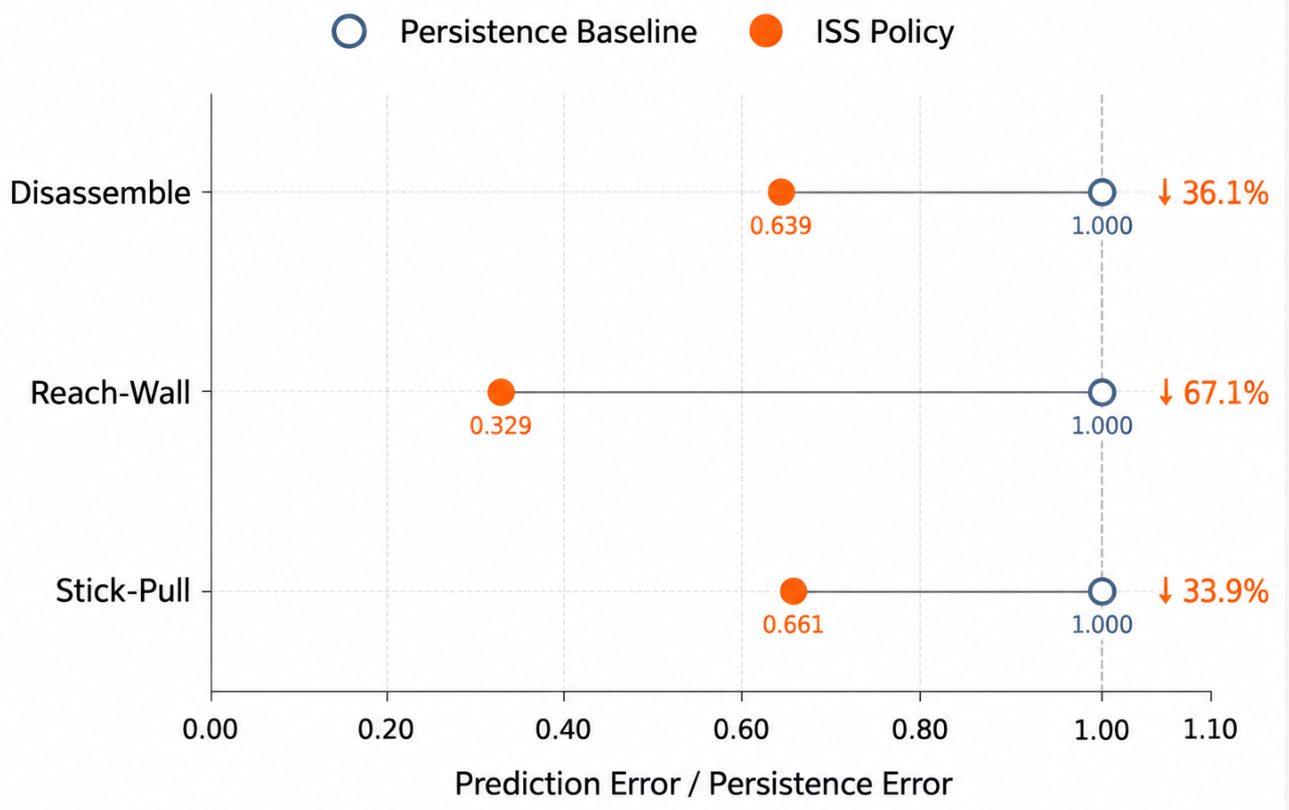}
        \par\smallskip
        {\footnotesize (b) Future scene prediction}
    \end{minipage}
    \caption{\textbf{Analysis of implicit scene supervision.}
    (a) Normalized robot motion prediction error compared with an action-and-state
    Ridge predictor. (b) Future point-cloud latent prediction error relative to
    the persistence baseline, which copies the current latent.}
    \label{fig:representation_analysis}
\end{figure}

To further examine the predictive information learned by the auxiliary modules, we analyze checkpoints on Disassemble, Reach-Wall, and Stick-Pull. Figure~\ref{fig:representation_analysis} (left) shows that our motion predictor achieves lower errors than the action-and-state Ridge baseline. On the right, ISS Policy predicts future scene features more accurately than copying the current latent, supporting the effectiveness of the auxiliary supervision.

\begin{table}[tbp]
\centering
\caption{\textbf{Ablation on ISS loss weight $\lambda_{\mathrm{ISS}}$.} We compare different ISS loss weights across four tasks.}
\label{tab:iss_weight_ablation}
% resizebox 会自动调整字体大小以适应宽度
\resizebox{0.99\linewidth}{!}{%
    \setlength{\tabcolsep}{4pt} % 稍微增加列间距，因为去除了竖线，阅读更轻松
    \begin{tabular}{c|cccc|c}
    \toprule
    $\lambda_{\mathrm{ISS}}$ & Disassemble & Reach-Wall & Stick-Pull& Pick-Place-Wall & Average\\
    \midrule
    0.2 & 87          & 82          & 82          & 85          & 84.00 \\
    0.4 & \textbf{93} & \textbf{85} & \textbf{90} & 91          & \textbf{89.75} \\
    0.6 & 92          & 83          & 82          & \textbf{93} & 87.50 \\
    0.8 & 91          & 77          & 87          & 89          & 86.00 \\
    1.0 & \textbf{93} & 80          & 78          & 90          & 85.25 \\
    \bottomrule
    \end{tabular}%
}
\end{table}

\noindent\textbf{Impact of ISS Weight.}
The ISS weight $\lambda_{\mathrm{ISS}}$ multiplies the combined auxiliary loss in Eq.~\eqref{eq:total_loss}, whereas $\lambda_m$, $\lambda_s$, and $\lambda_c$ weight its individual components. As shown in Table~\ref{tab:iss_weight_ablation}, the performance peaks at $\lambda_{\mathrm{ISS}}=0.4$, achieving the highest average success rate of \textbf{89.75\%}. While increasing the weight from 0.2 to 0.4 improves the results, further increasing it beyond this point leads to a gradual decline in performance. This indicates that a moderate weight strikes the optimal balance, providing sufficient geometric supervision to structure the feature space without interfering with the primary policy learning objective.
%%%%%%%%%%%%%%%%%%%%%%%%%%%%%%%%%%%%%%%%%%%%%%%%%%%%%%%%%%%%%%%%%%%%%%%%%%%

%%%%%%%%%%%%%%%%%%%%%%%%%%%%%%%%%%%%%%%%%%%%%%%%%%%%%%%%%%%%

% \noindent\textbf{Ablations on Horizon Length.}

\begin{figure}[tbp]
    \centering
    \includegraphics[width=\linewidth]{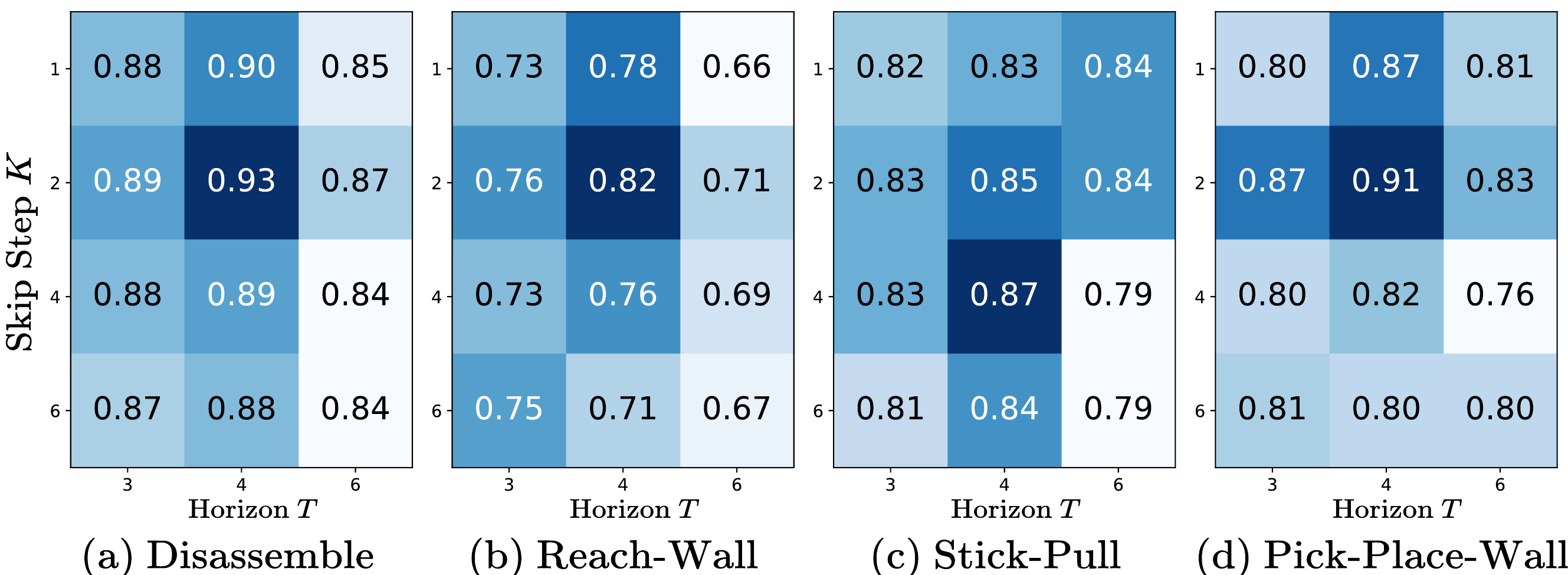}
    \caption{\textbf{Ablation on Skip Step $K$.} We compare skip step with different prediction horizon across four tasks.}
    \label{fig:heatmap}
\end{figure}

\noindent\textbf{Impact of Different Skip Steps.}
As shown in Fig.~\ref{fig:heatmap}, ISS Policy achieves highest performance under moderate skip lengths and prediction horizons. Compared to short-step prediction with $K=1$, using a moderate skip ($K=2\sim4$) encourages the ISS head to forecast more pronounced geometric changes, yielding richer geometric supervision and boosting success rates by roughly 2$\sim$5 percentages. In contrast, when the skip becomes too large (e.g., $K=6$) or the prediction horizon is overly extended (e.g., $T=6$), performance consistently degrades across tasks, suggesting that predicting scenes that are too far into the future makes the auxiliary task unnecessarily difficult and noisy, thereby weakening its beneficial effect on the action decoder.\looseness=-1

%%%%%%%%%%%%%%%%%%%%%%

%%%%%%%%%%%%%%%%%%%%%%% 

\begin{figure}[tbp]
    \centering
    \includegraphics[width=0.9\linewidth]{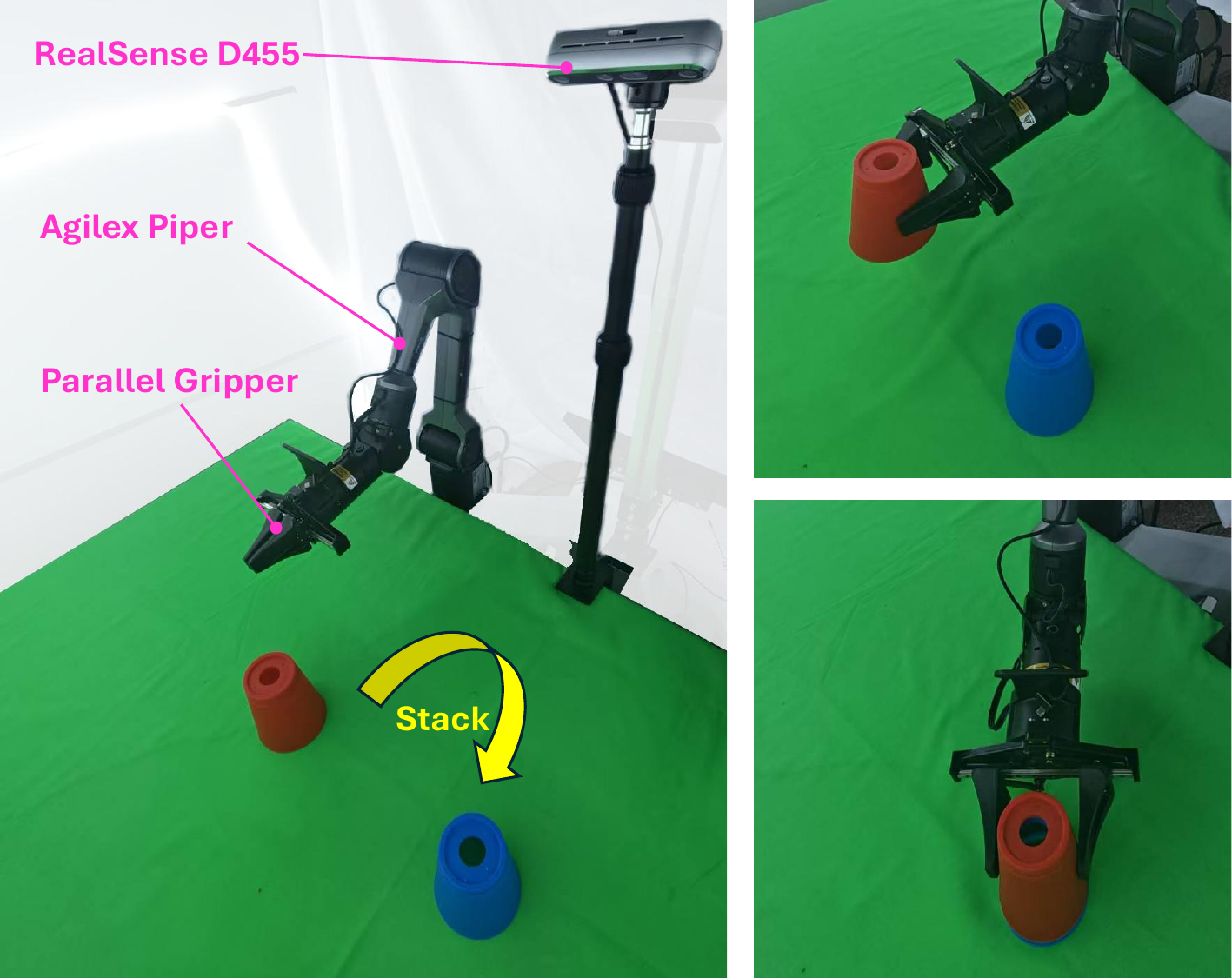}
    \caption{\textbf{Realworld Experiment Setup and Single-arm experiment.} }
    \label{fig:task}
\end{figure}

\begin{table}[tbp]
\centering
\caption{Realworld performance comparison over dual-arm tasks : Bottle Grasping, Cup Placing, Block Stacking.}
\label{tab:realworld}
\begin{tabular}{lcccc}
\toprule
Method & BG & CP & BS & Inference Time \\
\midrule
DP3 & 6/10& 2/10& 4/10 & 161.4\,ms \\
ISS Policy & 8/10 & 5/10 & 7/10 & 52.7\,ms \\
\bottomrule
\end{tabular}
\end{table}

\begin{figure}[tbp]
    \centering
    \includegraphics[width=0.95\linewidth]{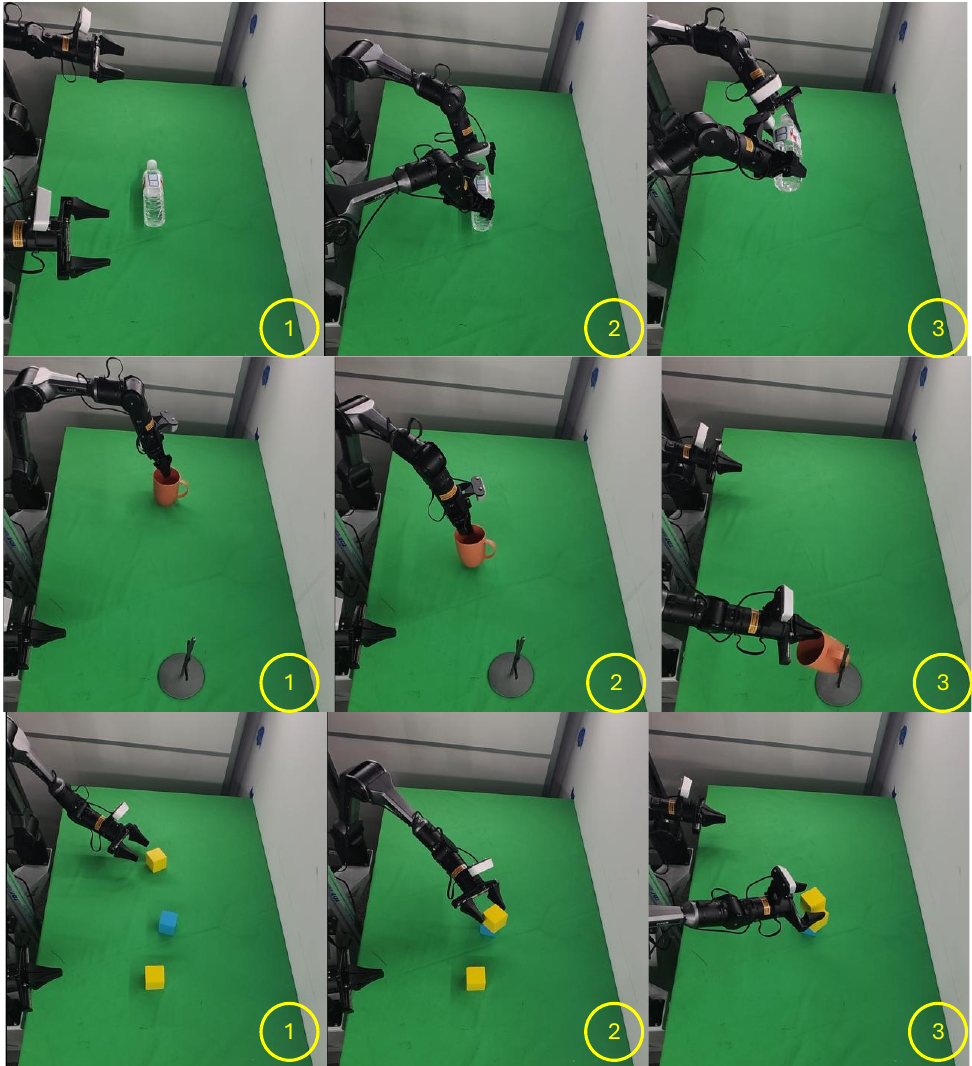}
    \caption{\textbf{Realworld Dual-arm Experiments.} 
    % The image sequence illustrates the robot successfully grasping and stacking the cups, validating the effectiveness of ISS Policy in real-world manipulation tasks.
    }
    \label{fig:realworld}
\end{figure}

\subsection{Real-World Experiments}

\noindent\textbf{Experiment Settings.}
We evaluate ISS Policy on three real-world dual-arm manipulation tasks using two AgileX Piper robot arms: bottle grasping, cup placing, and block stacking. A single Intel RealSense D455 RGB-D camera is used to capture visual observations, which are converted into point-cloud inputs for the policy. The policy is deployed on an NVIDIA RTX 4060 GPU for on-robot inference. The real-world setup and task procedures are shown in Fig.~\ref{fig:task}. For each task, we collect 50 teleoperated expert demonstrations for training. During evaluation, each policy is tested for 10 trials on each task under the same experimental setup.

%%%%%%%%%%%%%%%%%%%%%%

%%%%%%%%%%%%%%%%%%%%%%% 

\noindent\textbf{Results.} 
As shown in Figs.~\ref{fig:task} and~\ref{fig:realworld}, ISS Policy is evaluated on three real-world dual-arm manipulation tasks: bottle grasping, cup placing, and block stacking. These tasks involve different object geometries and bimanual coordination requirements, allowing us to examine the real-world applicability of the proposed method in dual-arm manipulation scenarios. The quantitative results are summarized in Table~\ref{tab:realworld}. The reported success rates are computed over 10 evaluation trials for each policy on each task, and the average success rate is obtained across all three tasks.

%%%%%%%%%%%%%%%%%%%%%%

%%%%%%%%%%%%%%%%%%%%%%% 

\FloatBarrier
\section{CONCLUSIONS}

We presented \textbf{ISS Policy}, a scalable 3D diffusion policy connecting action generation to future scene prediction through supervised robot motion prediction. Experiments demonstrate improved simulation and real-world performance, while scaling studies show gains from larger models and more demonstrations. Component ablations support the effectiveness of conditional fusion, implicit scene supervision, and scheduled sampling. Crucially, the auxiliary predictors are disabled at deployment, making implicit scene supervision inference-cost-free while bringing improved policy performance and learning efficiency.

% \addtolength{\textheight}{-12cm}   

%%%%%%%%%%%%%%%%%%%%%%%%%%%%%%%%%%%%%%%%%%%%%%%%%%%%%%%%%%%%%%%%%%%%%%%%%%%%%%%%

%%%%%%%%%%%%%%%%%%%%%%%%%%%%%%%%%%%%%%%%%%%%%%%%%%%%%%%%%%%%%%%%%%%%%%%%%%%%%%%%



% \section*{ACKNOWLEDGMENT}

% The preferred spelling of the word ÒacknowledgmentÓ in America is without an ÒeÓ after the ÒgÓ. Avoid the stilted expression, ÒOne of us (R. B. G.) thanks . . .Ó  Instead, try ÒR. B. G. thanksÓ. Put sponsor acknowledgments in the unnumbered footnote on the first page.

%%%%%%%%%%%%%%%%%%%%%%%%%%%%%%%%%%%%%%%%%%%%%%%%%%%%%%%%%%%%%%%%%%%%%%%%%%%%%%%%

% Keep the bibliography together, after all experimental figures.
\clearpage
\balance
\bibliographystyle{IEEEtran}    % 这里用你上传的 .bst 名字（不带 .bst 后缀）
\bibliography{IEEEfull}         % 这里是你的 .bib 文件名（不带 .bib 后缀）

\end{document}